\documentclass[preprint,5p]{elsarticle}

\usepackage{hyperref}

\journal{IEEE Transactions on Computational Social Systems}

\usepackage[caption=false]{subfig}
\usepackage{tabularx}
\usepackage{adjustbox}
\usepackage{caption}
\usepackage{dblfloatfix}
\usepackage{footmisc}

\begin{document}

\begin{frontmatter}

\title{Bias-Aware Face Mask Detection Dataset}


\author[1]{Alperen Kantarcı\corref{cor1} }
\ead{kantarcia@itu.edu.tr}
\author[2]{Ferda Ofli}
\ead{fofli@hbku.edu.qa}
\author[2]{Muhammad Imran}
\ead{mimran@hbku.edu.qa}
\author[1]{Hazım Kemal Ekenel}
\ead{ekenel@itu.edu.tr
}
\cortext[cor1]{Corresponding author}
\address[1]{Department of Computer Engineering, Istanbul Technical University, Istanbul, Turkey}
\address[2]{Qatar Computing Research Institute, Hamad Bin Khalifa University, Doha, Qatar}





\begin{abstract}

In December 2019, a novel coronavirus (COVID-19) spread so quickly around the world that many countries had to set mandatory face mask rules in public areas to reduce the transmission of the virus. To monitor public adherence, researchers aimed to rapidly develop efficient systems that can detect faces with masks automatically. However, lack of representative and novel datasets proved to be the biggest challenge. Early attempts to collect face mask datasets did not account for potential race, gender, and age biases. Therefore, the resulting models show inherent biases toward specific race groups, such as Asian or Caucasian. In this work, we present a novel face mask detection dataset that contains images posted on Twitter during the pandemic from around the world. Unlike previous datasets, the proposed Bias-Aware Face Mask Detection (BAFMD) dataset contains more images from underrepresented race and age groups to mitigate the problem for the face mask detection task. We perform experiments to investigate potential biases in widely used face mask detection datasets and illustrate that the BAFMD dataset yields models with better performance and generalization ability. The dataset is publicly available at \url{https://github.com/Alpkant/BAFMD}.

\end{abstract}


\end{frontmatter}

\section{Introduction} \label{introduction}
The rapid worldwide spread of the severe acute respiratory syndrome coronavirus 2 (SARS-CoV2) or COVID-19 created a global pandemic. More than 127 million cases were confirmed within a year~\cite{who-cases} because of the virus. Medical experts, public health agencies, and governments worldwide recommended a series of prevention measures, such as social distancing, travel bans, country-wide lockdowns, and wearing face masks in public spaces~\cite{covid_measures}. 
Practical measures, such as face masks, have been adopted for more extended periods. Computer vision researchers and practitioners rapidly started developing automatic detection methods due to this massive increase in face mask usage, as existing face detection methods struggled to detect faces with masks. Since monitoring and screening applications of face mask detection systems help society prevent virus transmission, it became essential to develop an accurate and fair face mask detection system. 

\begin{figure*}[!t]
	\centering
	\includegraphics[width=\linewidth]{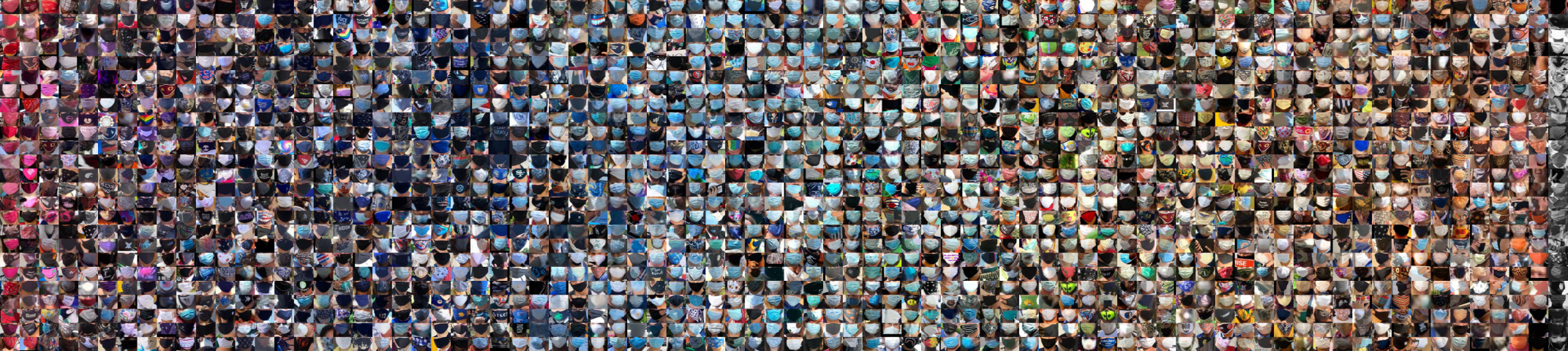}
	\caption{\label{fig:mask_visualization} [Best viewed in color] Example face mask images available in the proposed dataset. Unlike simulated or the pre-pandemic datasets, various colors and textures of the face masks are present.}
\end{figure*}

Face mask detection has been an understudied sub-topic within face detection research until the COVID-19 pandemic. Most early work on occluded face detection focused on occlusions such as glasses, hands covering the lower part of the face, and pollution-masks~\cite{mafa, occluded1, occluded2, occluded3}. Moreover, these early works only focused on Asian countries where face mask usage was already common even before the COVID-19 pandemic because of excessive air pollution and SARS-associated coronavirus~\cite{leduc2004sars}.
Therefore, when the pandemic started, researchers combined available datasets \cite{maskdetection_initial} that contained Asian people wearing face masks with other standard face detection datasets, such as WIDER~\cite{wider}, or they tried to produce datasets with artificial face masks. Although these approaches showed better performance for the masked face detection task, their application in the real-world setting remained limited mainly due to imbalanced race distribution in the datasets. 
Biased data leads to biased models that may not be applicable to certain population segments, e.g., people with dark skin color. Such issues potentially raise ethical concerns about the fairness of automated systems. Therefore, a system that will be used in daily life across the world should be trained with a more representative and demographically balanced dataset to mitigate biases \cite{fairface, biases_demogprahics}. A study~\cite{merler2019diversity} shows that most existing large-scale face databases are biased towards ``lighter skin'' faces, e.g., Caucasian, compared to ``darker'' faces, e.g., Black. However, such a study has not been conducted on face mask detection datasets. In our observations, we noticed a clear selection bias toward Asian faces, as the most famous face mask datasets are collected in Asia.


In this paper, we address the need for a representative face mask detection dataset with particular focus on the bias and fairness aspects of the problem. To this end, we propose a new dataset, which has a more balanced distribution across gender, race, and age, using images from Twitter around the world. We make the dataset publicly available\footnote{\label{footnote1}The dataset is available at \url{https://github.com/Alpkant/BAFMD}} to support future research. We summarize demographic statistics about dataset using publicly available state-of-the-art face attribute prediction methods. We experiment with existing face mask detectors as well as our newly proposed model on widely used face mask detection datasets. Finally, we show that our bias-aware dataset leads to models that can generalize and perform better than the state-of-the-art face mask detection models.

\section{Related Work} \label{relatedwork}

Face occlusion~\cite{robust_occlusion, face_occlusion, face_detection_occlusion}, object detection~\cite{yolov3, tan2020efficientdet}, and face detection~\cite{face_detection_rcnn, deng2020retinaface} are well-researched fields that can provide good baselines for developing face mask detection systems. However, face mask detection has received limited attention among the detection tasks and was studied within broader occluded face detection problem. Therefore, only a few datasets were available when the COVID-19 pandemic started. During the pandemic, researchers published numerous studies in the face mask detection field. These studies mostly focus on either collecting new datasets or combining different datasets to obtain a representative face mask detection dataset as listed in Table~\ref{tab:table1}. However, the high cost of annotating a new dataset prevented most of the researchers from collecting face mask datasets. Thus, researchers focused on either creating artificial face masks on face images~\cite{maskedface_artificial} or refining the annotations of the publicly available face occlusion datasets~\cite{fmld}. 
\begin{figure}[b]
	\centering
	\includegraphics[width=\linewidth]{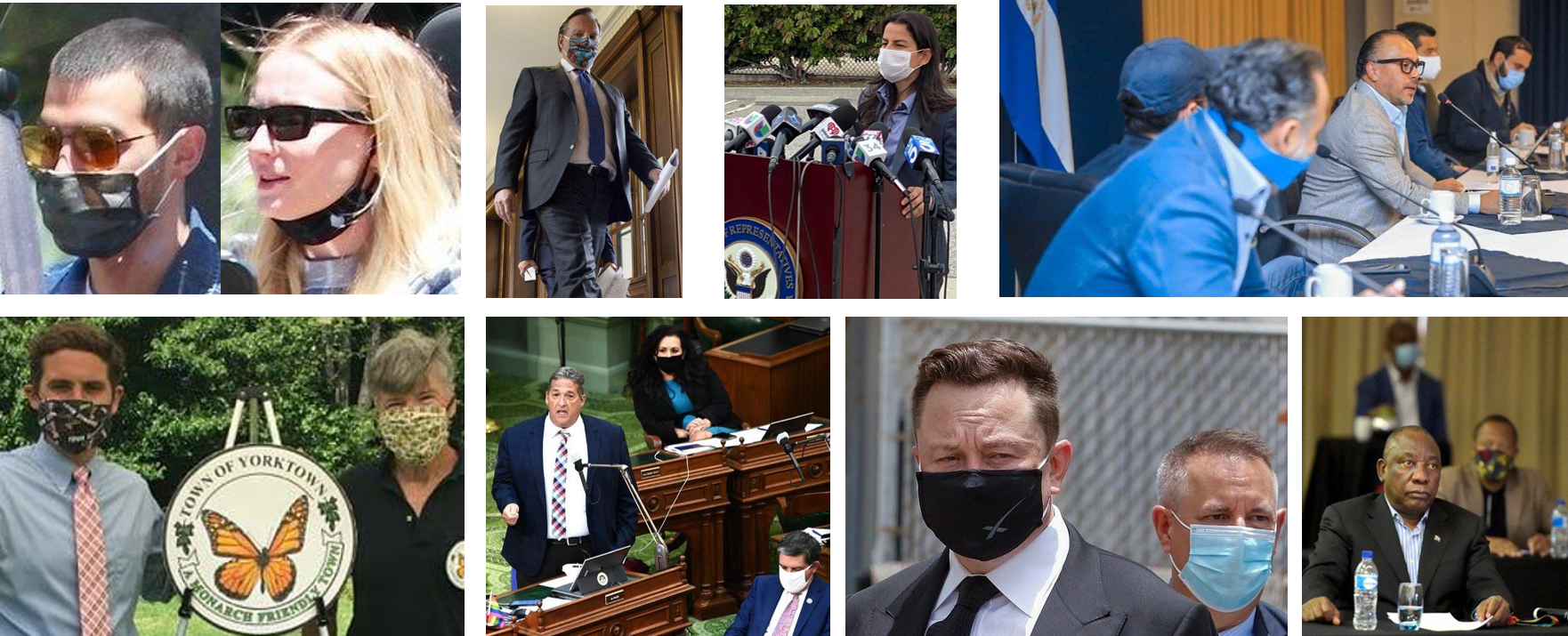}
	\caption{\label{fig:dataset_examples}Example images from the Bias-Aware Face Mask Detection (BAFMD) dataset.}
\end{figure}

\begin{table*}[h!]
\centering
\begin{tabular}{@{}l|c|c|c|c|c|c@{}} 
\textbf{Dataset name} & \textbf{\#Mask} & \textbf{\#No Mask} & \textbf{\#Images} & \textbf{Mask Type} & \textbf{Image Source} & \textbf{Ethnicity}   \\
    \hline
  MAFA~\cite{mafa} & 35,806 & 911 & 30,811 & Real & Google + Bing & Asian  \\
  FMD~\cite{face_mask_kaggle} & 3,232 & 840 & 853 & Real & Unknown & Asian \\
  MFDD~\cite{mfdd_masked} & 24,771 & Unkown & 4,343 & Real & \cite{aizootech} + Internet & Unknown  \\
  FMLD*~\cite{fmld} & 29,532 & 33,540 & 41,934 & Real &  MAFA + WIDER & Asian + Caucasian  \\
  MMD~\cite{mmd} & 6,758 & 2,309 & 6,024 & Real & Internet & Various \\
  MaskedFace-Net~\cite{maskedface_artificial} & 67,049 & 66,734 & 133,783 & Artificial & FFHQ~\cite{ffhq} & Various\\
  ISL-UFMD~\cite{eyiokur2021_dataset} & 10,698 & 10,618 & 21,816 & Real & Internet & Various\\
  \textbf{BAFMD (ours)} & 13,492 & 3,118 & 6,264 & Real & Twitter & Various  \\
\end{tabular}
\caption{We compare different face mask detection datasets which contain bounding box annotations for the detection task. (*) symbol indicates that the corresponding dataset only proposes annotations for existing datasets.}
\label{tab:table1}
\end{table*}

Previous works that proposed combination of datasets \cite{fmld,mfdd_masked} mostly use the MAFA dataset~\cite{mafa}, which was collected from the Internet in 2017 as a face occlusion detection dataset. MAFA contains various face occlusions, including face masks. However, most of the images are collected from Asian countries, where face masks are widely used by the population. This is also the case for many different face mask detection datasets, such as MFDD~\cite{mfdd_masked}. Having a racial bias in training dataset is a huge drawback for creating universal face mask detection models as they would be biased towards specific race groups. In contrast, we collected images from different ethnicities and age groups to create a more representative dataset. Furthermore, our dataset contains variety of face mask designs and textures, that increased during the COVID-19 pandemic. 
Fig.~\ref{fig:mask_visualization} visualizes the diverse nature of the face masks while Fig.~\ref{fig:dataset_examples} shows some sample images available in the proposed dataset. 
This way, our dataset ensures that trained face mask detection models are capable of detecting faces from different ethnicities and age groups with face masks of not only white and blue, as typically used in previous years, but of different colors and shapes. The MAFA dataset also contains many incorrect annotations, as shown in \cite{fmld}. Therefore, modifying the MAFA dataset to create a new face mask detection dataset requires fixing the incorrect annotations. 

One of the initial works on face mask datasets was presented in \cite{mfdd_masked} which proposed three different datasets for masked face recognition and face mask detection. The authors propose Masked-Face Detection Dataset (MFDD), which is the extended version of the MAFA and WIDER datasets for face mask detection. They also propose the Real-world Masked-Face Recognition Dataset (RMFRD) and the Simulated Masked-Face Recognition Dataset (SMFRD) for masked-face recognition. RMFRD contains frontal face images that are collected from the Internet, whereas in SMFRD, facial masks were added artificially to simulate masked faces. Unfortunately, only a subset of these datasets are publicly available. Furthermore, training models with simulated images can be problematic due to the high domain difference between real and artificial masks.

Another dataset that filters previously proposed datasets to create a more refined one is proposed in \cite{fmld}. Authors annotate the MAFA~\cite{mafa} and WIDER~\cite{wider} datasets in the context of the COVID-19 pandemic and with respect to placement-correctness of face mask, gender, ethnicity, and pose. All of these annotations are manually generated and provide coarse predictions of pose and ethnicity attributes. The authors also indicate the necessity of demographic attributes in face mask detection datasets. Their annotations show that the MAFA dataset contains mostly Asian and the WIDER dataset contains mostly Caucasian faces. This is  problematic, because the trained models might associate mask usage with races, as MAFA contains masked faces and WIDER mainly contains faces without masks.

Face Mask Detection (FMD) dataset~\cite{face_mask_kaggle} is proposed for a Kaggle competition during the pandemic. Images were collected from the Internet. They are annotated for three classes: with mask, without mask, and mask worn incorrectly. It contains 4072 face annotations of 853 images. Medical Mask Detection (MMD) dataset~\cite{mmd} has been acquired from the Internet with paying attention to the diversity of ethnicities, ages, and regions. All images have been manually curated and annotated. It covers 20 classes of different accessories including faces with a mask, without a mask, or with an incorrectly worn mask. 

MaskedFace-Net dataset~\cite{maskedface_artificial} is an artificially created dataset using a deformable mask model and facial landmarks, similar to SMFRD~\cite{mfdd_masked}. 
Face images are collected from Flickr-Faces-HQ~\cite{ffhq} (FFHQ) dataset. Then, digitally created mask models are placed on the mouth area of the given face images and annotated according to the correct mask usage. 

Finally, more recently, researchers  
collected images from publicly available face datasets (i.e., FFHQ~\cite{ffhq}, CelebA~\cite{celeba}, LFW~\cite{LFWTech}), YouTube, and web crawling from websites to create Interactive Systems Labs Unconstrained Face Mask Dataset (ISL-UFMD) \cite{eyiokur2021_dataset}. Having diverse and multiple sources of images naturally increase the variability of ethnicity, age, and gender within the dataset. Unlike ISL-UFMD, we quantitatively measure specific attributes of the faces to increase the diversity and reduce possible biases in our dataset in a systematic manner.

\section{Proposed Dataset} \label{dataset}

\begin{figure*}[b!]
\centering
\raisebox{73pt}{\parbox[c]{.05\textwidth}{BAFMD}}%
\subfloat[FairFace Gender Predictions][FairFace Gender Predictions]{
\includegraphics[width=0.3\textwidth]{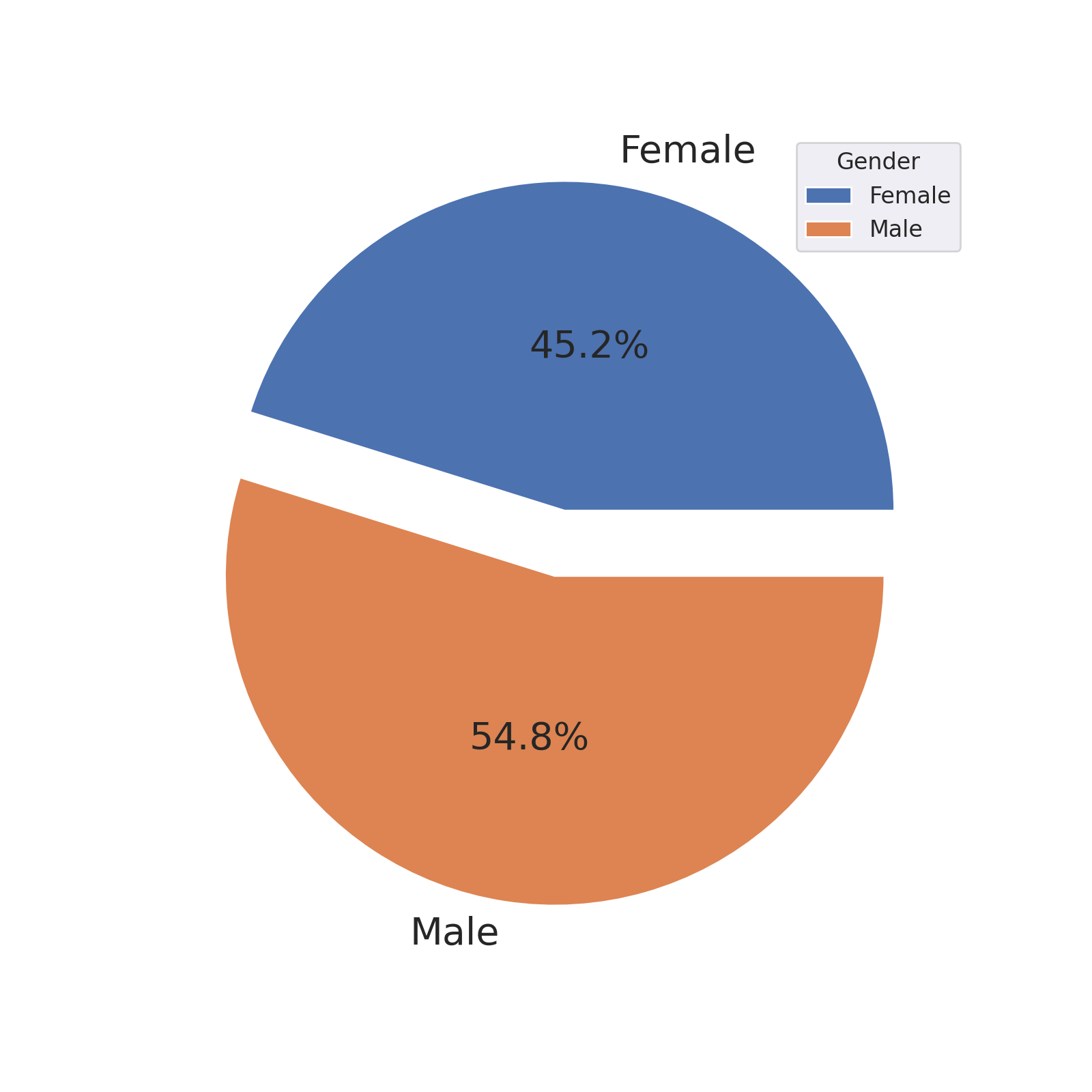}
\label{fig:statistics_subfig1}}
\subfloat[FairFace Race Predictions][FairFace Race Predictions]{
\includegraphics[width=0.3\textwidth]{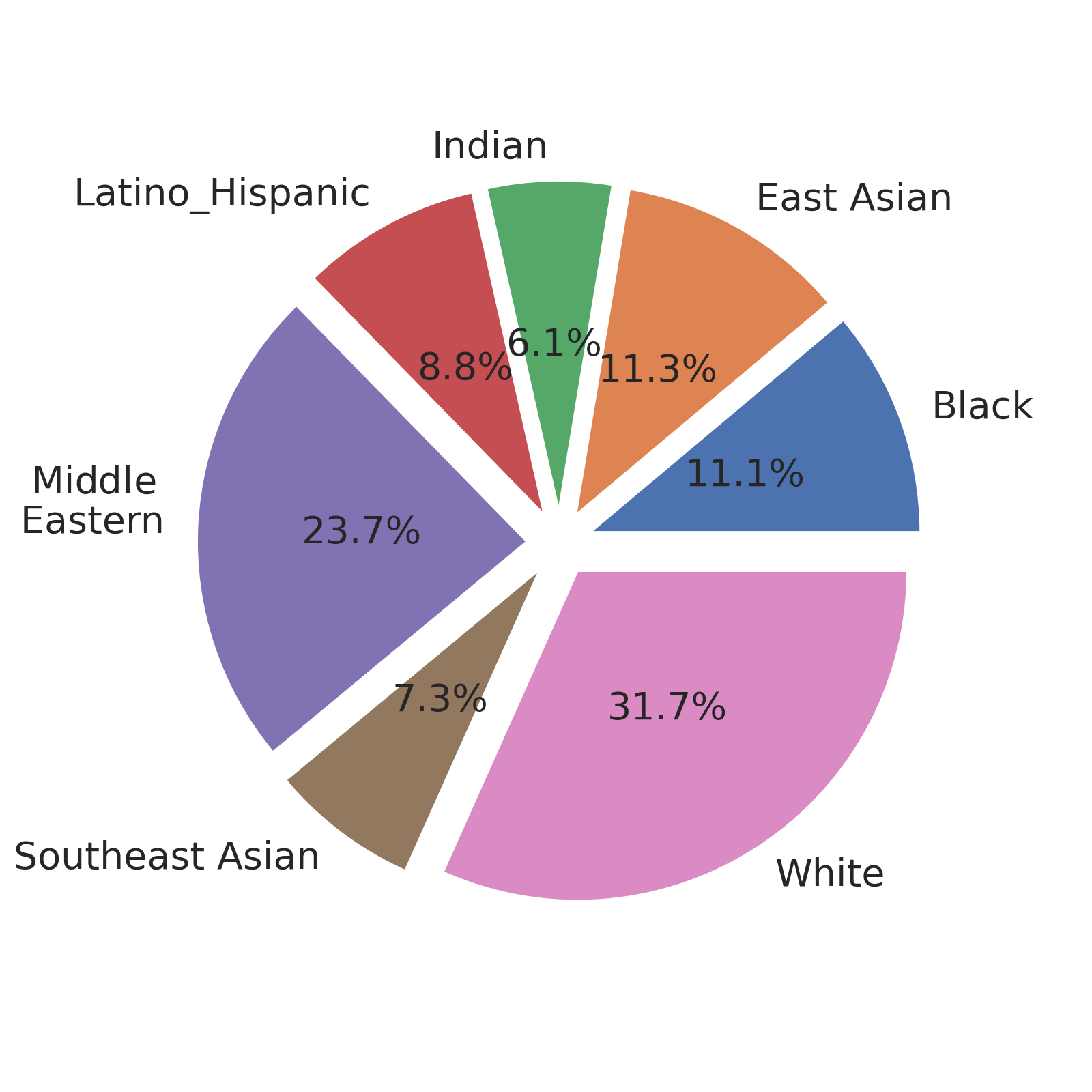}
\label{fig:statistics_subfig2}}
\subfloat[FairFace Age Predictions][FairFace Age Predictions]{
\includegraphics[width=0.3\textwidth]{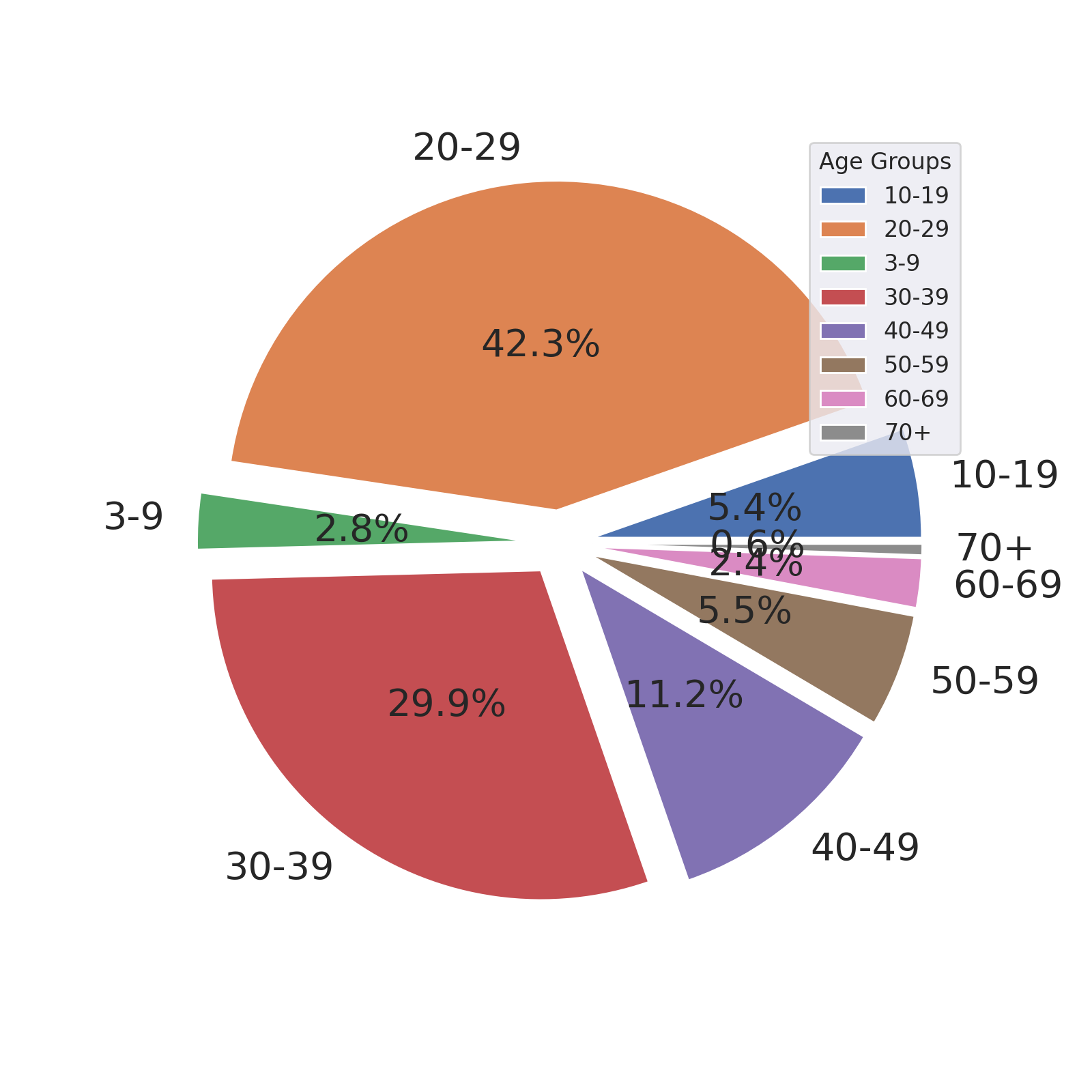}
\label{fig:statistics_subfig3}}
\par
\raisebox{73pt}{\parbox[c]{.05\textwidth}{MAFA}}%
\subfloat[FairFace Gender Predictions][FairFace Gender Predictions]{
\includegraphics[width=0.3\textwidth]{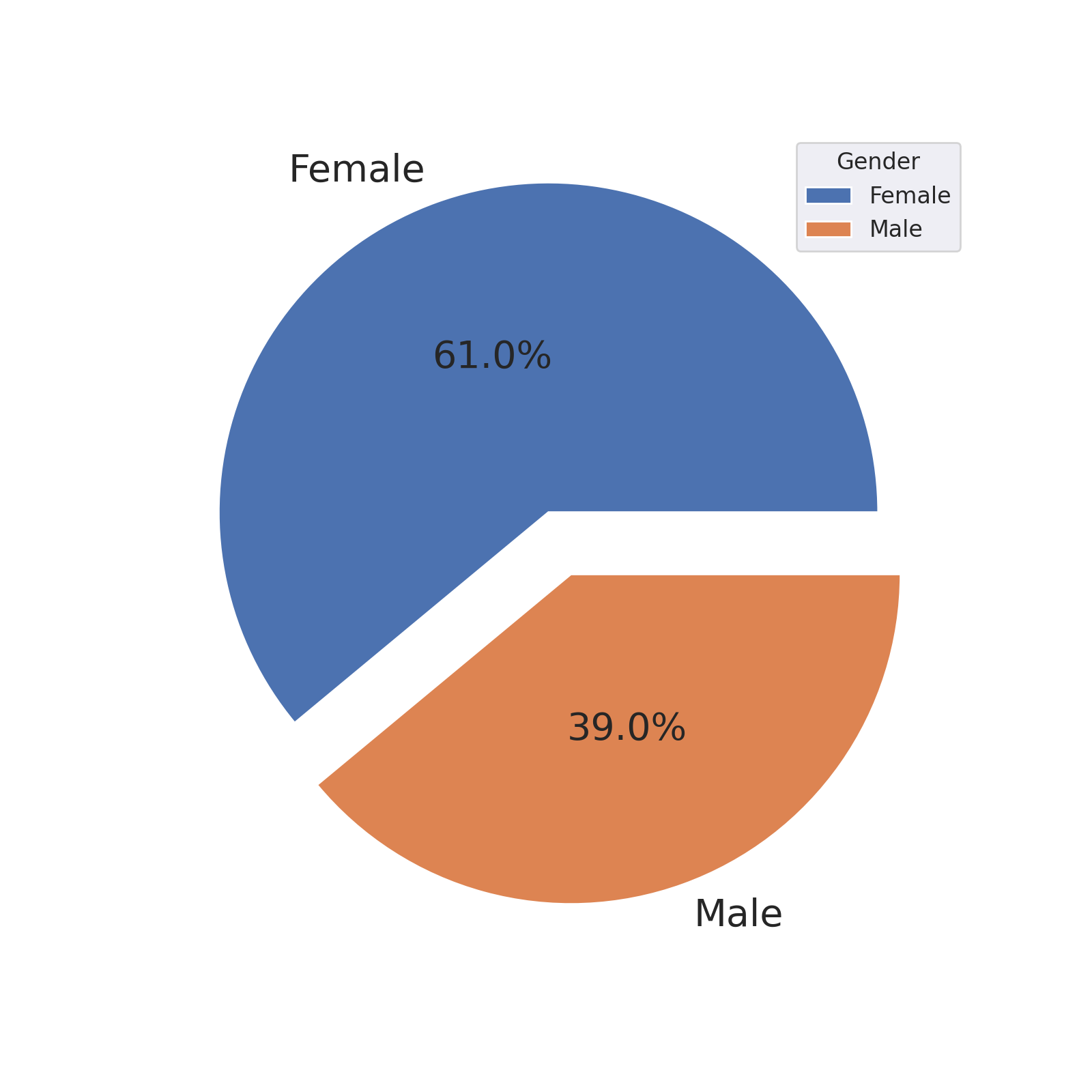}
\label{fig:statistics_subfig4}}
\subfloat[FairFace Race Predictions][FairFace Race Predictions]{
\includegraphics[width=0.3\textwidth]{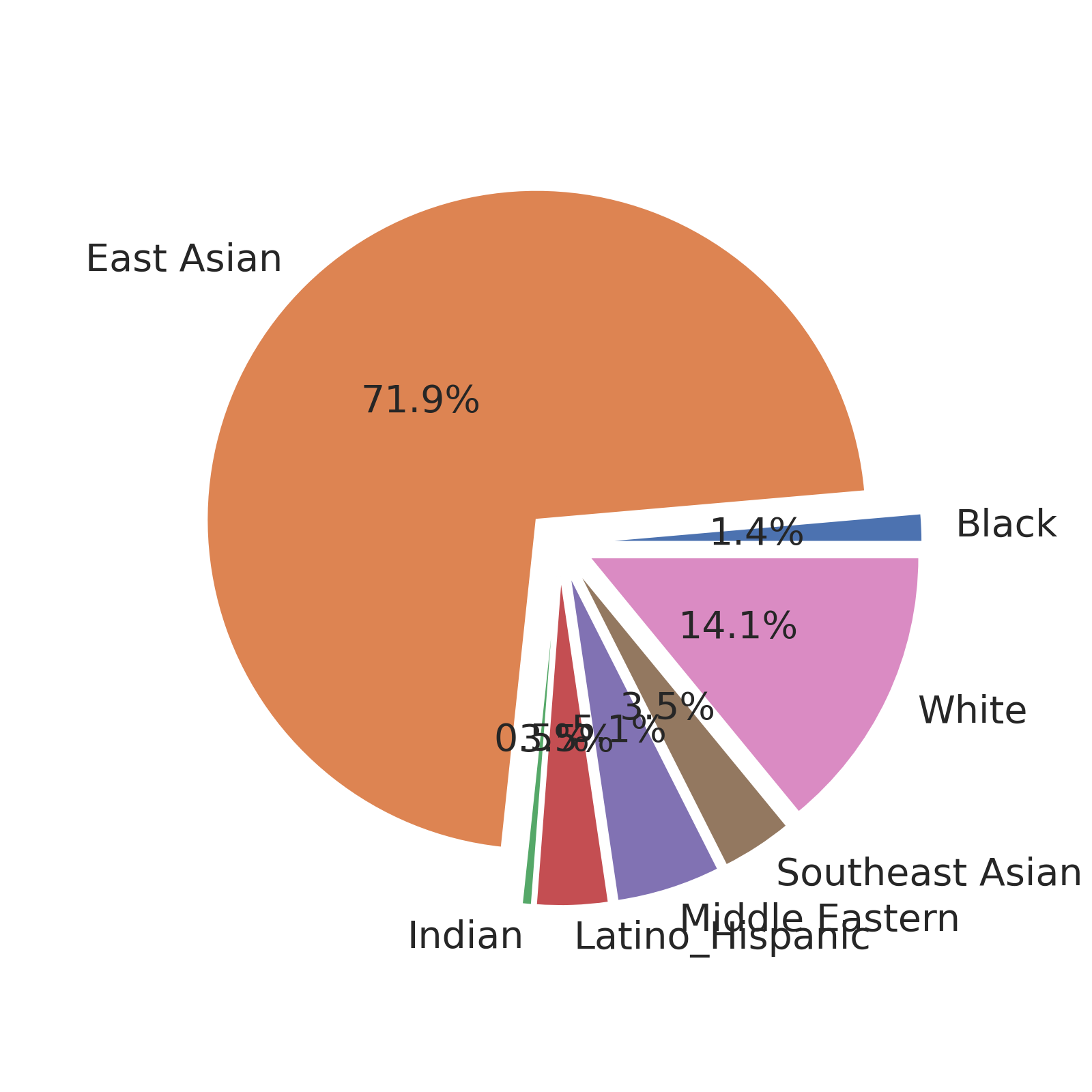}
\label{fig:statistics_subfig5}}
\subfloat[FairFace Age Predictions][FairFace Age Predictions]{
\includegraphics[width=0.3\textwidth]{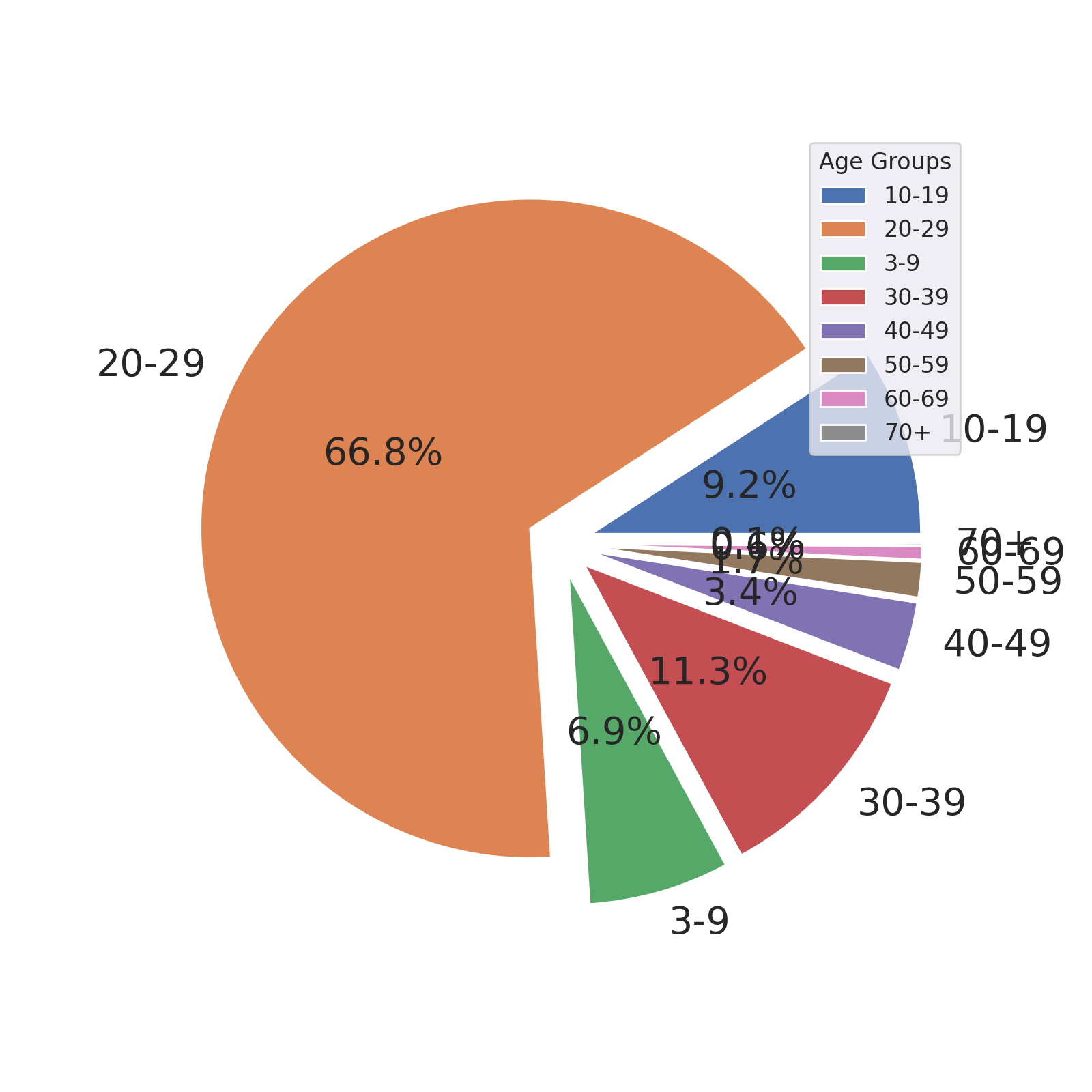}
\label{fig:statistics_subfig6}}
\caption{FairFace analysis tool pipeline has been executed over all images of the BAFMD and MAFA datasets. FairFace gender, race and age group predictions for TFMD dataset are presented in \ref{fig:statistics_subfig1}, \ref{fig:statistics_subfig2}, \ref{fig:statistics_subfig3}, respectively. Similarly, for MAFA dataset gender, race and age group predictions are presented in \ref{fig:statistics_subfig4}, \ref{fig:statistics_subfig5}, \ref{fig:statistics_subfig6}, respectively.}
\label{fig:dataset_statistics}
\end{figure*}

Race and gender biases are well-known but an understudied topic for the face mask detection task.
Our primary focus is to gather images that are as representative as possible to reduce dataset bias for a specific ethnicity, age, or gender. 
To this end, we first collected publicly posted images from Twitter by using keywords related to COVID-19 prevention measures and face masks during the pandemic. Tweet collection was initially restricted to Los Angeles County as it is the second most diverse place in the United States according to the Racial and Ethnic Diversity Index of Census Bureau~\cite{losangelesdiversity}. Therefore, it is a suitable location to obtain a diverse collection. We ran a state-of-the-art (SOTA) face detector~\cite{deng2020retinaface} to eliminate images without faces. Then, we manually labeled faces with and without masks by annotating facial bounding box locations and mask usage. We used LabelImg~\cite{labelimg} labeling tool for annotations. By using this manually labeled data, we trained YOLO-v5 \cite{yolov5}, which is a SOTA object detection model. The trained YOLO-v5 model is utilized to speed up our data annotation process by adopting a semi-automatic label annotation pipeline to estimate candidate bounding boxes and class labels.

For developing a representative and demographically balanced face mask dataset, having a balanced ratio of different faces is important. In our image collection, we created race, age group, and gender predictions of people. 
We employed FairFace~\cite{fairface}, which is a SOTA face attribute classifier trained on a balanced race and gender face attribute dataset.
%
FairFace requires MTCNN~\cite{mtcnn} face detector due to its training pipeline. Therefore, we only produced predictions for faces that could be detected with MTCNN~\cite{mtcnn}. FairFace defines seven race groups: White, Black, Indian, East Asian, Southeast Asian, Middle Eastern, and Latino. It also has an option to define five race groups by combining Middle Eastern and White, as well as East Asian and Southeast Asian. Aside from race predictions, we also used gender and age group predictions. FairFace~\cite{fairface} uses the following age groups: 0-2, 3-9, 10-19, 20-29, 30-39, 40-49, 50-59, 60-69, and 70+. To balance the racial distribution and get more images from underrepresented ethnicities, we expanded our location filter to include images from 56 different countries, such as Kenya, Canada, Vietnam, and Turkey. 
Final dataset comprises 6,264 images which contain 13,492 faces with masks and 3,118 faces without masks. Unlike most previous face mask detection datasets, which contain only one face per image, the high number of faces indicates that our dataset also contains crowded scenes. Moreover, our dataset captures high pose and illumination variations. Fig.~\ref{fig:dataset_examples} shows images from our dataset, which we named as Bias-Aware Face Mask Detection (BAFMD) dataset.

We compare our dataset with the MAFA~\cite{mafa} dataset, a well-known and widely used face mask detection dataset. Specifically, we compare the ratios of race, gender, and age groups. Having a more balanced dataset in terms of race, gender, and age groups creates less bias for the trained models~\cite{fairface}. 
As illustrated in Fig.~\ref{fig:dataset_statistics}, our dataset achieves more balanced ratios across race, gender, and age groups.

For reproducibility, we define training and testing sets of the dataset\textsuperscript{\ref{footnote1}}. To create a test set, we used the statistics given race predictions of the FairFace model. We kept the test set proportional to the racial, gender, and age group ratios. We used 25\% of the faces as the testing set. In the end, we got 5,466 training images and 798 testing images with a similar racial distribution. As stated above, FairFace~\cite{fairface} requires face images cropped by MTCNN~\cite{mtcnn}. Therefore, in order to produce reliable predictions from FairFace, we use MTCNN on our dataset to crop images. As MTCNN cannot detect all masked faces, only a subset of the dataset can be used for facial attribute prediction. We assume this subset would be sufficient to give information about the entire dataset.

\section{Masked Face Detection} \label{method}
In this work, we use a state-of-the-art object detection architecture, YOLO-v5~\cite{yolov5}, for training a facial mask detection model. Multiple research analyzed the performance of different single- and two-stage object detection models on face mask detection datasets. Moreover, many studies in the field investigated face detection and classification using two separate networks. In this work, we compare six face and face mask detection models. We propose to use YOLO-v5 model as a face mask detector and compare the YOLO-v5 model with five different state-of-the-art face and face mask detectors. YOLO-v5 model is an extension to YOLO-v3~\cite{yolov3} model. YOLO object detectors divide images into a grid system. Each cell in the grid is responsible for detecting objects within itself. A single forward pass of the model yields multiple bounding boxes and their class prediction probabilities. Therefore, they provide faster and better object detection results compared to the other object detectors. YOLO-v5 contains multiple new features over YOLO-v3, such as Path aggregation network~\cite{panet} and Cross Stage Partial Network~\cite{cspnet}. 

In classical object detection, millions of images are annotated; therefore, bigger models like YOLO-v5 Extra Large can be trained. We train the YOLO-v5 Small model due to limited number of images in face mask detection datasets, and initialize training with the pretrained model weights. 
For each experiment, we start with a learning rate of 0.001 and use the learning rate scheduler of YOLO-v5. We train each model up to 450 epochs with an early stopping criterion to avoid overfitting.

For comparing YOLO-v5 model with the state-of-the-art face mask detectors, we use MTCNN~\cite{mtcnn}, Baidu~\cite{baidu}, AIZooTech~\cite{aizootech}, RetinaFace~\cite{deng2020retinaface}, and AntiCov~\cite{anticov}. For all of the detectors, we use the default hyperparameters proposed in their paper or code. MTCNN~\cite{mtcnn} is one of the most popular and successful face detector which consists of three cascaded neural networks. Baidu~\cite{baidu} detector is based on PyramidBox~\cite{baidu} single-shot face detector. PyramidBox~\cite{baidu} implements several strategies to use context information to improve the face detection results. AIZooTech~\cite{aizootech} is one of the first proposed face mask detection networks. It is a single-shot detector customized for the face mask detection problem. RetinaFace~\cite{deng2020retinaface} is a single-shot multi-level face localization method that performs pixel-wise face localization. We use RetinaFace model with ResNet-50~\cite{resnet} backbone network. The AntiCov~\cite{anticov} is a customized one-stage face detector based on RetinaFace~\cite{deng2020retinaface}. The AntiCov is much faster and lighter than RetinaFace in order to deploy the model on end devices with limited computation power.

\section{Experiments} \label{experiments}
In this section, we first present the metrics used to assess the performance of the face mask detection methods. Then, we conduct experiments to evaluate the performance of different face mask detection methods on widely used face mask detection datasets and our BAFMD dataset. Additionally, we test different face mask detection methods on different datasets while changing the training dataset to assess the representativeness, i.e. generalization capability, of the training datasets. Finally, we consider the risks of using social media images where the contents can be removed in time. To observe the effect of this phenomenon, we test the performance of our model with respective to the changing number of training images.    
In all experiments, we use standard object detection performance metric, mean average precision (mAP), which has been proposed in \cite{pascalvoc} and adopted with different object detection benchmarks \cite{coco}. Calculation of the mAP requires the computation of the Intersection over Union (IoU) for each class. We calculate
IoU by using area of our prediction ($P$) and area of ground truth ($G$) bounding box for an object.
Following the most common object detection competitions, we consider a prediction as \textit{True Positive} (TP) if its IoU score is greater than $0.5$, i.e.\ $mAP_{0.5}$.



\subsection{Same-Dataset Experiments}
\begin{table}[t]
\centering
\begin{tabular}{@{}l|cccc@{}}
   & \multicolumn{4}{c}{\textbf{Dataset}}                                 \\ \hline
\textbf{Method}   & \textbf{MAFA}          & \textbf{WIDER}         & \textbf{FMLD}          & \textbf{BAFMD}          \\ \hline
MTCNN~\cite{mtcnn}              & 42.5          & 85.6          & 65.8          & 34.3          \\
Baidu~\cite{baidu}              & 59.4          & 88.5          & 77.2          & 58.7          \\
AIZooTech~\cite{aizootech}          & 85.1          & 89.3          & 86.5          & 76.4          \\
RetinaFace~\cite{deng2020retinaface}         & 81.2          & \textbf{99.4} & 91.9 & 73.6          \\
AntiCov~\cite{anticov} & 84.9          & 93.7          & 87.8          & 78.1          \\
Ours               & \textbf{87.3} & 92.0          & \textbf{92.2}          & \textbf{86.8}
\end{tabular}

\caption{Mean Average precision ($mAP_{0.5}\%$) results of different face detection models on MAFA~\cite{mafa}, WIDER~\cite{wider}, FMLD~\cite{fmld} and BAFMD datasets. WIDER face is a well known face detection dataset and does not include any mask annotation. Other datasets contain both mask and no mask classes. Please note that FMLD~\cite{fmld} dataset is combination of MAFA~\cite{mafa} and WIDER~\cite{wider} datasets. Model that performed best on each dataset is highlighted in bold.}
\label{tab:mafa_wider_rfmd_train}
\end{table}
A considerable amount of face mask detection models are trained with combination of MAFA and WIDER datasets because they contain high number of images and were readily available at the start of the COVID-19 pandemic. 
However, as MAFA dataset included some noisy annotations, combining MAFA and WIDER dataset required more work. In FMLD dataset~\cite{fmld}, the authors proposed a combination of MAFA and WIDER dataset by annotating both datasets manually. Therefore, they created a better dataset for training face mask detectors. In order to be comparable with previous work, we use MAFA, WIDER, FMLD and our Bias-Aware Face Mask Detection (BAFMD) dataset. In our experiments, we compare our model against MTCNN \cite{mtcnn}, Baidu \cite{baidu}, AIZooTech \cite{aizootech}, and RetinaFace-AntiCov \cite{anticov}. As explained in Sections \ref{relatedwork} and \ref{method}, these models and datasets are widely used for face mask detection.

The results in Table~\ref{tab:mafa_wider_rfmd_train} show that in MAFA dataset most of the proposed detectors achieve 80 to 85 $mAP_{0.5}\%$. However, in WIDER dataset, the performances range from 85 to 99 $mAP_{0.5}\%$. This is an indication for difficulty of face mask detection problem. As FMLD dataset is a combination of both MAFA and WIDER, the performances of models on this dataset are higher than MAFA but lower than WIDER. In our proposed dataset, the performances of different models are slightly worse than MAFA dataset, which implies the difficulty of the dataset. Our proposed YOLO-v5 model outperforms other detectors on three out of four datasets. Moreover, performance of the YOLO-v5 model is more stable across different datasets than other detectors.

\subsection{Cross-Dataset Experiments}
Many widely-used, publicly available face mask detection datasets are racially imbalanced and contain images from specific regions of the world, such as Asia. In order to create a better and more representative dataset, we collected images from all around the world while keeping a balanced racial distribution. To test the representativeness of the datasets, we train RetinaFace~\cite{deng2020retinaface} and our proposed method on FMLD and BAFMD datasets, seperately. We chose FMLD dataset as it combines MAFA and WIDER face datasets which are among the popular datasets on face detection and face mask detection. 
For both datasets we use their standard training and testing sets. We used the same hyperparameters as in the within-dataset experiments. Table~\ref{tab:interdataset} shows that the performance of both RetinaFace and our model decrease when trained on one dataset and tested on another. When models are trained on FMLD and tested on BAFMD, the drop in $mAP_{0.5}\%$ is nearly 15\%. On the other hand, when models are trained on BAFMD and tested on FMLD, the drop in $mAP_{0.5}\%$ is nearly 7\%. This experiment shows that a more representative and racially balanced dataset, such as BAFMD, can lead to better generalization. Therefore, using BAFMD may serve as a better training set for general face mask detectors. Apart from the better performance, training with a balanced dataset enable models to have less accuracy discrepancy among all race and gender groups as shown in FairFace study~\cite{fairface}.
\begin{table}[t]
\centering
\begin{tabular}{c|c|c|c} 
\textbf{Method} & Training Set & Test Set & $mAP_{0.5}\%$ \\
\hline
RetinaFace & BAFMD & BAFMD & 73.6\\
RetinaFace & FMLD & FMLD & 91.9\\
RetinaFace & BAFMD & FMLD & 84.0\\
RetinaFace & FMLD & BAFMD & 60.2\\
\hline
Ours & BAFMD & BAFMD & 86.9\\
Ours & FMLD & FMLD & 92.2\\
Ours & BAFMD & FMLD & 84.5\\
Ours & FMLD & BAFMD & 72.9\\
\end{tabular}
\caption{RetinaFace~\cite{deng2020retinaface} and Our method have been trained on both FMLD~\cite{fmld} and BAFMD datasets to assess their performance on a dataset that have not been trained. First four rows show the performance of RetinaFace~\cite{deng2020retinaface} model when trained and tested on different sets. On the other hand last four rows show the performance of our model in the same settings. We also show the same-dataset test performances to highlight the performance drop on cross-dataset tests. }
\label{tab:interdataset}
\end{table}

\subsection{Robustness to Volatile Social Media Data}

Everyday, social media users share thousands of photos to express their ideas or show what is happening around them. In many social media platforms users can control with whom to share their content. For example, a user can share their photo publicly and then can make it private so that only the people that they allow can see. Moreover, users can delete or edit their shared content anytime. Therefore, social media content constantly changes and acquisition and processing of this content should also adopt to this changing environment. As our proposed dataset contains images from Twitter, we can not expect to retrieve the entire dataset completely as time passes and the number of samples that can be accessed through the shared links is likely to decrease by time.

In order to assess the performance of our models against removal of data in time, we trained different face mask detection models using fractions of the same training and validation sets of our BAFMD dataset. 
Six experiments were held by using 30\%, 40\%, 50\%, 60\%, 80\%, and 100\% of all training and validation samples, while the test set is kept fixed to be able to assess the performance fairly. The removed samples were chosen randomly in order to maintain a consistent distribution across different splits. This experimental setup indicates the potential performance drop for the researchers who would like to develop a face mask detection system using BAFMD dataset. 
\begin{table}[t]
\centering
\begin{tabular}{c|c} 
\textbf{Percentage of Training Images} & \textbf{$mAP_{0.5}\%$} \\
\hline
  100\% & \textbf{86.88}\\ 
  80\% & 84.12\\ 
  60\% & 82.46\\ 
  50\% & 81.52\\ 
  40\% & 80.75\\ 
  30\% & 79.20\\ 
\end{tabular}
\caption{For each training we keep randomly selected images of training and validation sets. First column shows percentage of images that has been kept for training to the original size of the dataset. 
}
\label{tab:removedata}
\end{table}

For face mask detection, we used our proposed YOLO-v5~\cite{yolov5} model. In order to make the comparisons fair, we used the same hyperparameters for all the trainings. 
In Table~\ref{tab:removedata}, we show performance of our models with respect to different amount of training data. When all of the available data is used for the training 86.9\% $mAP_{0.5}$ is achieved. Removing 10\% of the training images drops the performance by 1\% to 2\% in terms of $mAP_{0.5}$. Therefore, the results indicate that a small percentage of the dataset can still provide sufficient amount of information to train a successful face mask detector. 

\section{Conclusions} \label{conclusions}

We studied the problem of face mask detection during the COVID-19 pandemic with particular focus on dataset bias. Face mask detection problem has been an understudied sub-problem of face and object detection. In order to help society during the COVID-19 pandemic, many researchers across the world rapidly focused on the problem. However, majority of the earlier work has simply focused on training new architectures with the limited number of face occlusion datasets.

In this work, we introduced a novel face mask detection dataset named as Bias-Aware Face Mask Detection (BAFMD) dataset. To the best of our knowledge, it is the first face mask detection dataset that has been collected with a focus on mitigating demographic bias. Unlike most publicly available datasets, our dataset contains real-world face mask images with a more balanced distribution across different demographics, e.g., gender, race and age.

Moreover, our experimental results on multiple publicly available datasets show that the proposed model has comparable or superior performance to the proposed methods for face mask detection. 
We demonstrated that YOLO-v5 can be a good model candidate for face mask detection problem due to its low latency and superior performance.

\bibliography{references}

\end{document}